# COVID-19 PREDICTION BASED ON CT-SCANS OF LUNGS USING DENSENET ARCHITECTURE[1]

Deborup Sanyal
*Department of Computer Science and Engineering,
University at Buffalo*

Mingchen Gao, PhD
*Department of Computer Science and Engineering,
University at Buffalo*

---

[1] May 2022

**INTRODUCTION AND AIM:**

COVID-19 took the world by storm since December 2019. A highly infectious communicable disease, COVID-19 is caused by the SARS-CoV-2 virus. By March 2020, the World Health Organisation (WHO) declared COVID-19 as a global pandemic. A pandemic in the 21$^{st}$ century after almost 100 years was something the world was not prepared for which resulted in the deaths of around 1.6 million people worldwide. The most common symptoms of COVID-19 was associated with the respiratory system and resembled a cold, flu or pneumonia. After extensive research, doctors and scientists came to the conclusion that the main reason for lives being lost due to COVID-19 was failure of the respiratory system. Patients were dying gasping for breath. Top healthcare systems of the world was failing badly as there was an acute shortage of hospital beds, oxygen cylinders and ventilators. There were hundreds who were dying without getting any treatment at all.

Our aim for this project is to help doctors decide the severity of COVID-19 by reading the patient's Computed Tomography Scans or CT Scans of the lungs. Computer models are usually less prone to human errors and Machine Learning or Neural Network models tend to give better accuracy as the training on the dataset keeps getting better. We have decided on using a Convolutional Neural Network Model. Given that a patient tests positive, we want our model to analyse the severity of COVID-19 infection within 1 month of the positive test result. The severity of the infection might be promising or unfavourable (if it leads to intubation or death) completely based on the CT scans in the dataset that we will be using.

**DATASET AND DATA PRE-PROCESSING:**

We have decided on using the STOIC2021 dataset which consists of hundreds of CT scans from different patients with respect to the reverse transcription–polymerase chain reaction (RT-PCR) reference standard. CT scans are bound to early reports, RT-PCR, demographic information (age, weight, health) and clinical symptoms. The RT-PCR results are given in binary form.

The dataset contains .mha files. We can read the .mha files either by using MedPy or SimpleITK. MedPy is a Python library and script collection that provides basic functionality for reading, writing, and modifying large images of any dimensions. SimpleITK is an image analysis toolkit that includes a wide number of components for general filtering, image segmentation, and registration. We have thought of going forward with SimpleITK because even with MedPy, we had to extract ITK before working with it.

Our training set consists of 10,000 images which we trained all together for the prospect of this project. In our project for Advanced Machine Learning (CSE 674), we divided our dataset into 3 parts of 3,330 images each for easy understanding of results. Our validation set consists of 300 images.

Data pre-processing is a data mining technique which is used to transform the raw data in a useful and efficient format. Fig 1 denotes the various steps of data pre-processing that we used.

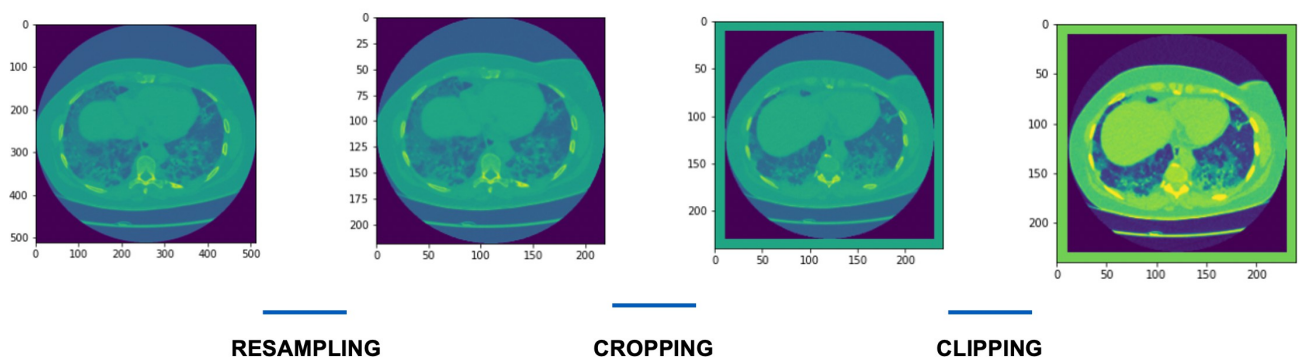

RESAMPLING                CROPPING                CLIPPING

Fig 1: Data Pre-processing

Resampling – Original dimensions of the images is 512 X 512 pixels. By resampling the images, we are bringing down the dimensions of the images to 224 X 224 pixels. A smaller dimension would mean taking up lesser space on the hard drive but still maintaining the same physical size.

Cropping – Cropping is one of the most basic photo manipulation processes, and it is carried out to remove an unwanted object or irrelevant noise from the periphery of a photograph, to change its aspect ratio, or to improve the overall composition. By cropping, we intend to focus on the main subject of the image while ignoring all other information (noise) as much as possible.

Clipping – In clipping, we tend to contrast the subject with the background so as to get a better visualization of the image.

**METHODS:**

Since usage of normal classifiers like Random Forest Classifier, Support Vector Machines (SVMs) etc. will not suffice the aim of this study due to the usage of image data, we have decided to use Convolutional Neural Networks, popularly known as CNNs. Reasons for using CNN are stated below –

- CNNs build their own features from raw signal. Opposed to other algorithms that use vector representations where every component usually makes some sense on its own. Pixels don't have meaning outside the context, but together they may contain more information about the object on a picture than a bunch of its properties that you feed into SVM.
- CNN is an excellent feature extractor, therefore utilizing it to classify medical images can avoid complicated and expensive feature engineering.
- CNNs can use infinitely strong priors and rely on spatial features. This is a primary property of max pooling layers. Good generalization and invariance to local fluctuations makes them super scalable.
- Multiple convolutional filters work and scan the complete feature matrix and carry out the dimensionality reduction.
- CNNs automatically detect the important features with minimum human supervision.
- CNNs are computationally efficient for medical image data.
- CNNs rely on spatial features. It can be their strength if the context of the feature is local (a bunch of pixels).

**EXPERIMENTAL LOGIC AND ARCHITECTURE:**

We used DenseNet as the base architecture to prepare our model. We had multiple reasons to choose DenseNet as our primary architecture which are listed as below:

- DenseNet has shown excellent accuracy despite having a fewer number of parameters. It is like ResNet as when the CNN becomes deep, the gradients appear to vanish.
- DenseNets improve gradient propagation by connecting all layers with each other. For example, a typical network with L layers has L connections (connection b/w the layers), however in DenseNet there will be L (L + 1) / 2 connections.
- Input layer creates K0 feature maps and then the first layer creates K feature maps and so on.
- As we go deeper into the network, the subsequent layers take input not only from the previous layer but all preceding layers including the input layer.
- The output of each of these layers is fixed. As we go deeper into the network, this becomes unsustainable. For example, let's say we have 10 layers, so the $10^{th}$ layer will take all the feature maps from the preceding 9 layers as input and now if each of these layers produce 128 or 256 feature maps, there is a feature map explosion. So to overcome this problem we have a fixed number of output maps from each layer.
- Each dense block contains a prespecified amount of layers in it. Output from a dense plot is given to a transition layer, which uses a 1 by 1 convolution followed by max pooling to reduce the size of the feature maps.
- DenseNet 121 has 121 layers.
- One of the main advantages of DenseNet is Parameter Efficiency that is, only a limited number of parameters/kernels are learnt per layer. Another important advantage of DenseNet is Feature Reuse that is, feature maps in all layers have direct access to the loss function and its gradient.
- Growth Rate is also an important part of DenseNet.
- Operations performed inside layer of DenseBlock can be summarised as–

$$\text{Batch Normalization} \rightarrow \text{ReLU} \rightarrow \text{Convolution}$$

- DenseNet on ImageNet : A DenseNet-BC structure with 4 dense blocks on 224 X 224 input images is used. The initial convolution layer comprises of 2k convolutions of size 7 X 7 with stride 2. The number of feature-maps in all other layers also follow from setting k.

| Layers | Output Size | DenseNet-121 | DenseNet-169 | DenseNet-201 | DenseNet-264 |
|---|---|---|---|---|---|
| Convolution | 112 × 112 | 7 × 7 conv, stride 2 | | | |
| Pooling | 56 × 56 | 3 × 3 max pool, stride 2 | | | |
| Dense Block (1) | 56 × 56 | $\begin{bmatrix} 1 \times 1 \text{ conv} \\ 3 \times 3 \text{ conv} \end{bmatrix} \times 6$ | $\begin{bmatrix} 1 \times 1 \text{ conv} \\ 3 \times 3 \text{ conv} \end{bmatrix} \times 6$ | $\begin{bmatrix} 1 \times 1 \text{ conv} \\ 3 \times 3 \text{ conv} \end{bmatrix} \times 6$ | $\begin{bmatrix} 1 \times 1 \text{ conv} \\ 3 \times 3 \text{ conv} \end{bmatrix} \times 6$ |
| Transition Layer (1) | 56 × 56 | 1 × 1 conv | | | |
| | 28 × 28 | 2 × 2 average pool, stride 2 | | | |
| Dense Block (2) | 28 × 28 | $\begin{bmatrix} 1 \times 1 \text{ conv} \\ 3 \times 3 \text{ conv} \end{bmatrix} \times 12$ | $\begin{bmatrix} 1 \times 1 \text{ conv} \\ 3 \times 3 \text{ conv} \end{bmatrix} \times 12$ | $\begin{bmatrix} 1 \times 1 \text{ conv} \\ 3 \times 3 \text{ conv} \end{bmatrix} \times 12$ | $\begin{bmatrix} 1 \times 1 \text{ conv} \\ 3 \times 3 \text{ conv} \end{bmatrix} \times 12$ |
| Transition Layer (2) | 28 × 28 | 1 × 1 conv | | | |
| | 14 × 14 | 2 × 2 average pool, stride 2 | | | |
| Dense Block (3) | 14 × 14 | $\begin{bmatrix} 1 \times 1 \text{ conv} \\ 3 \times 3 \text{ conv} \end{bmatrix} \times 24$ | $\begin{bmatrix} 1 \times 1 \text{ conv} \\ 3 \times 3 \text{ conv} \end{bmatrix} \times 32$ | $\begin{bmatrix} 1 \times 1 \text{ conv} \\ 3 \times 3 \text{ conv} \end{bmatrix} \times 48$ | $\begin{bmatrix} 1 \times 1 \text{ conv} \\ 3 \times 3 \text{ conv} \end{bmatrix} \times 64$ |
| Transition Layer (3) | 14 × 14 | 1 × 1 conv | | | |
| | 7 × 7 | 2 × 2 average pool, stride 2 | | | |
| Dense Block (4) | 7 × 7 | $\begin{bmatrix} 1 \times 1 \text{ conv} \\ 3 \times 3 \text{ conv} \end{bmatrix} \times 16$ | $\begin{bmatrix} 1 \times 1 \text{ conv} \\ 3 \times 3 \text{ conv} \end{bmatrix} \times 32$ | $\begin{bmatrix} 1 \times 1 \text{ conv} \\ 3 \times 3 \text{ conv} \end{bmatrix} \times 32$ | $\begin{bmatrix} 1 \times 1 \text{ conv} \\ 3 \times 3 \text{ conv} \end{bmatrix} \times 48$ |
| Classification Layer | 1 × 1 | 7 × 7 global average pool | | | |
| | | 1000D fully-connected, softmax | | | |

*Fig 2: Different DenseNet Architectures tested on the famous ImageNet dataset*

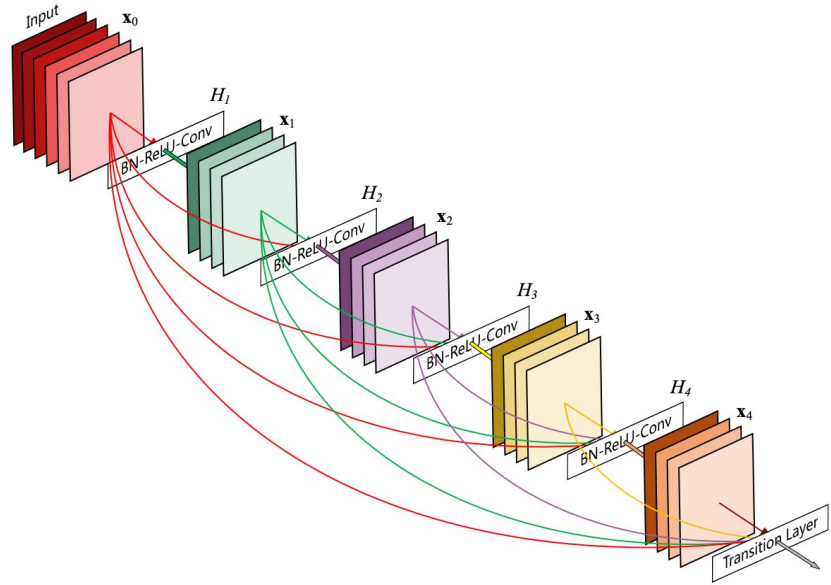

*Fig 3: The input layer feeds information to the Dense Block. Dense Block consists of n no. of layers which feeds to the transition layer. All layers take input from every preceding layer before it. Every layer in the dense block has Batch normalisation with ReLU activation function and then convolution.*

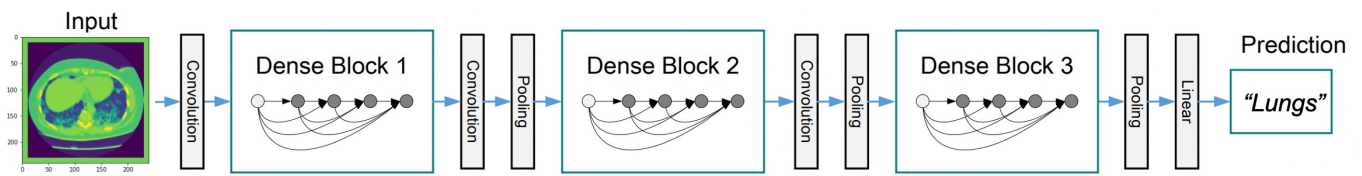

*Fig 4: DenseNet visualisation of our model using DenseNet121*

### DENSENET-169 vs DENSENET-121:

We had used DenseNet121 only for our project in Advanced Machine Learning. For our project this time, we have also tried using DenseNet169. Each architecture is made up of four DenseBlocks with variable levels. DenseNet-121 has [6,12,24,16] layers in the four dense blocks, whereas DenseNet-169 has [6, 12, 32, 32] layers. The main difference between DenseNet-121 and DenseNet-169 is the multiplier used by the DenseBlock.

**PROCEDURE:**

Our training dataset consists of 10,000 images. We did our training on the entire dataset at once instead of breaking the dataset into three phases. The procedure followed is as follows:

- We have the dataset given in the format of .mha files and a reference file with all the patient IDs along with the labels, Prob-Covid (which states the probability of having covid) and Prob-Severe (which states the severity of COVID-19).
- We used SITK Library for reading and pre-processing of the all the .mha files of the Patient IDs given in the reference file.
- We extracted the labels and for each Patient ID, we combined the pre-processed image as the input. The labels were given as the output in an iterable format for each Patient ID. We then created a PyTorch dataset.
- We used Sklearn to split our dataset into training dataset and validation dataset. We then created PyTorch datasets for both.
- Next, we used DataLoader of PyTorch because managing PyTorch dataset with DataLoader keeps the data manageable and helps to simplify the machine learning pipeline. DataLoader wraps an iterable around the dataset to enable easy access to the samples. We made DataLoader for both the training and validation datasets.
- From MonAi library, we used the DenseNet 121 and DenseNet169 model as explained before.
- We set the loss function to Pytorch.BCEWithLogitsLoss. This can be used with the raw output directly.
- We have used the Adam Optimizer with learning rate = 0.01
- We iterated over the Train DataLoader and trained our model for 100 epochs. (We did for 50 epochs in CSE 674.)
- For checking the accuracy, we have used both the labels together as an array (covid and severity), and then compared the array. We consider correct classification only if the final prediction array is same as the label.

**RESULTS:**

- **DENSENET-169 –** The results for DenseNet-169 are as below. The accuracy while evaluating the entire dataset for 100 epochs with DenseNet-169 resulted in 63.25%.

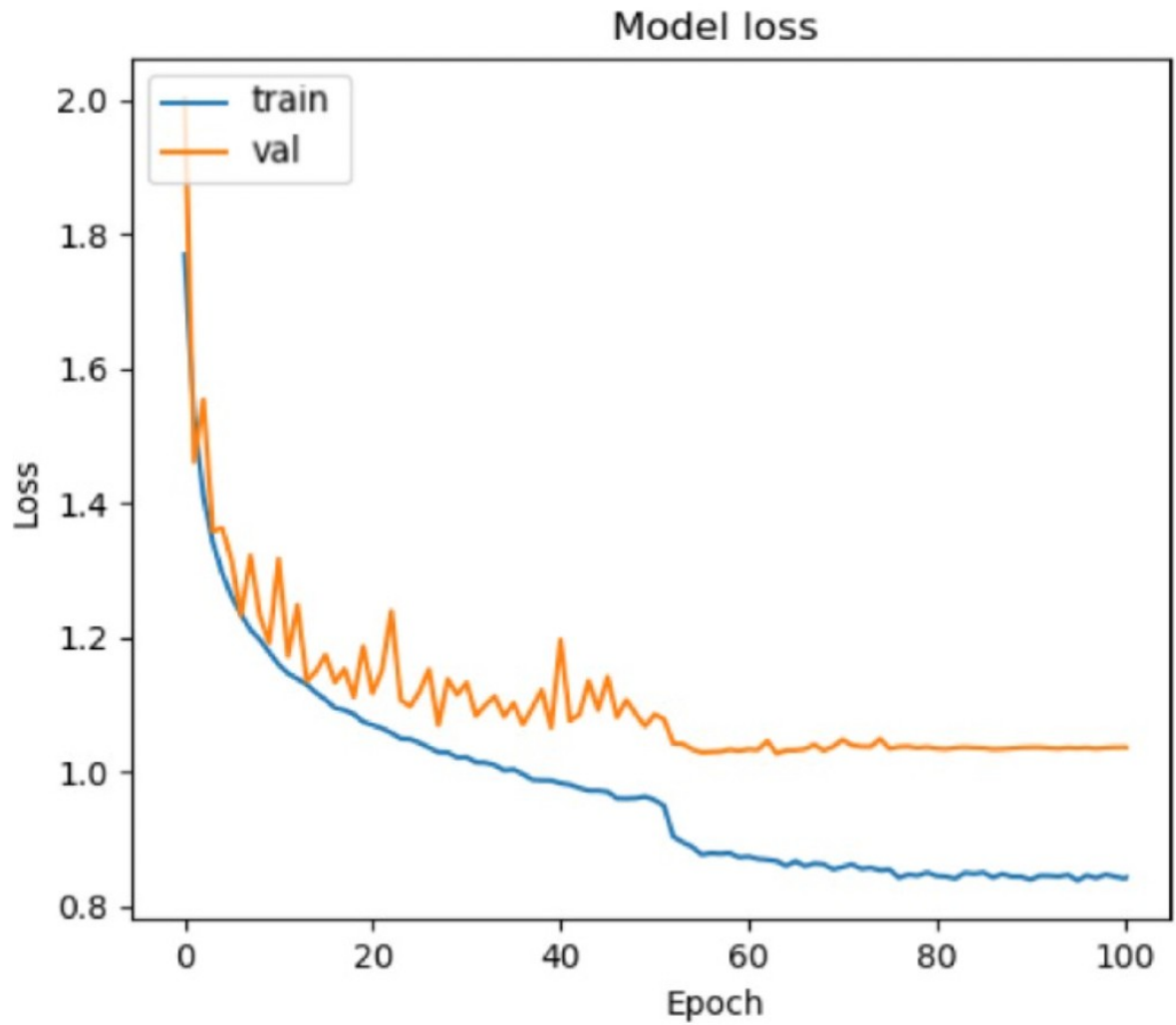

Fig: Loss graph for DenseNet-169

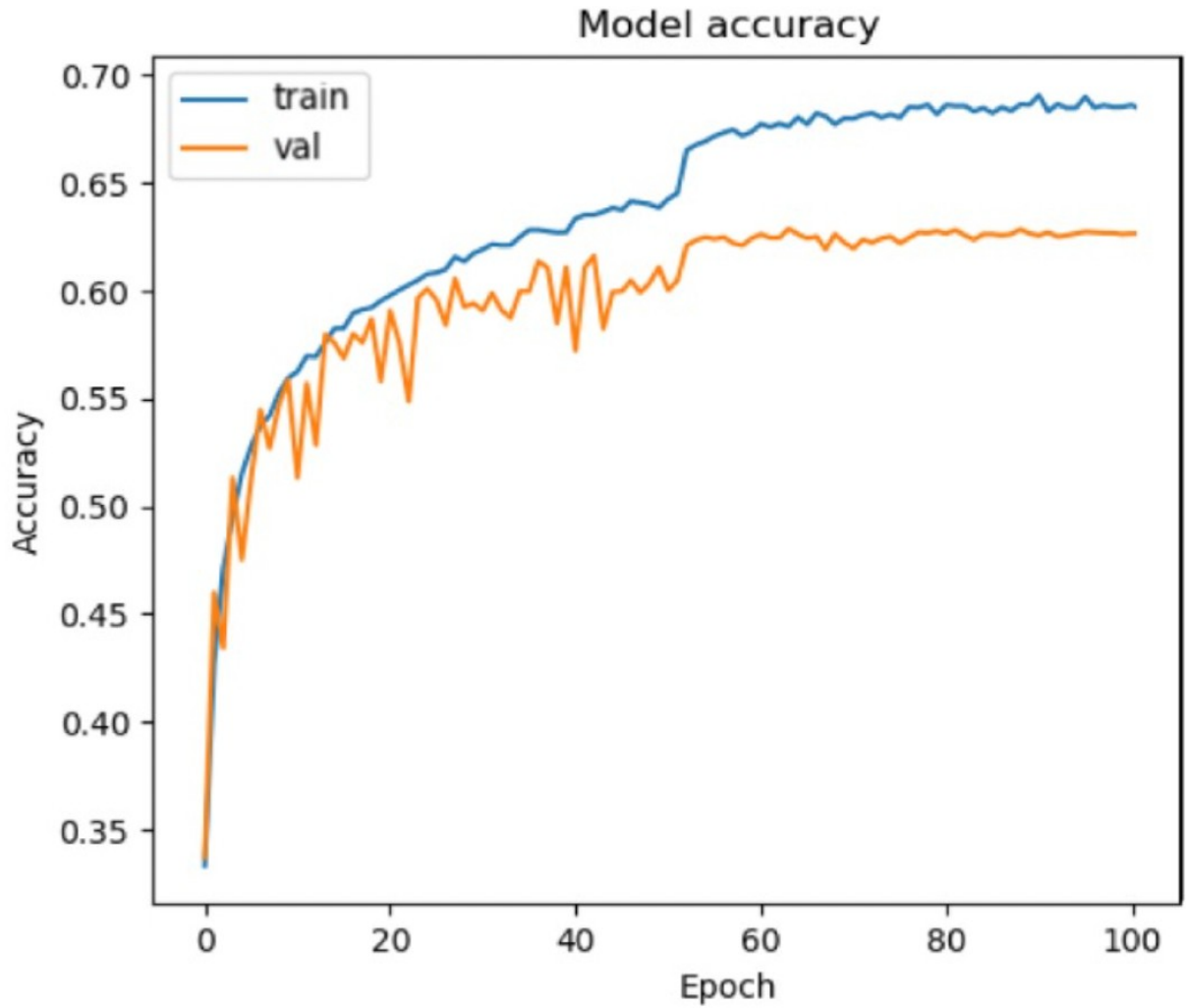

*Fig: Accuracy graph for DenseNet-169*

- **DENSENET-121 –** The results for DenseNet-121 are as below. The accuracy while evaluating the entire dataset for 100 epochs with DenseNet-121 resulted in 75.58%.

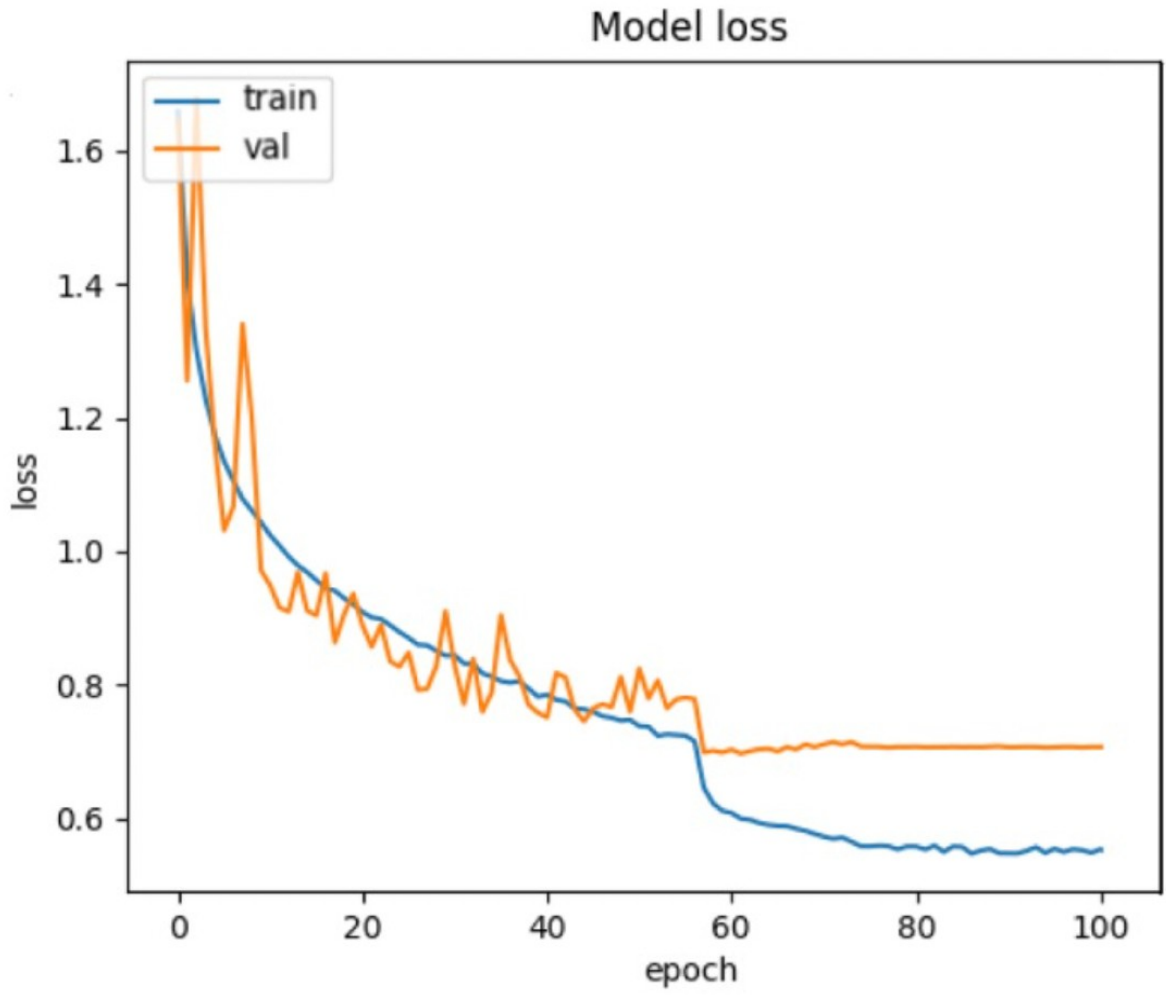

*Fig: Loss graph for DenseNet-121*

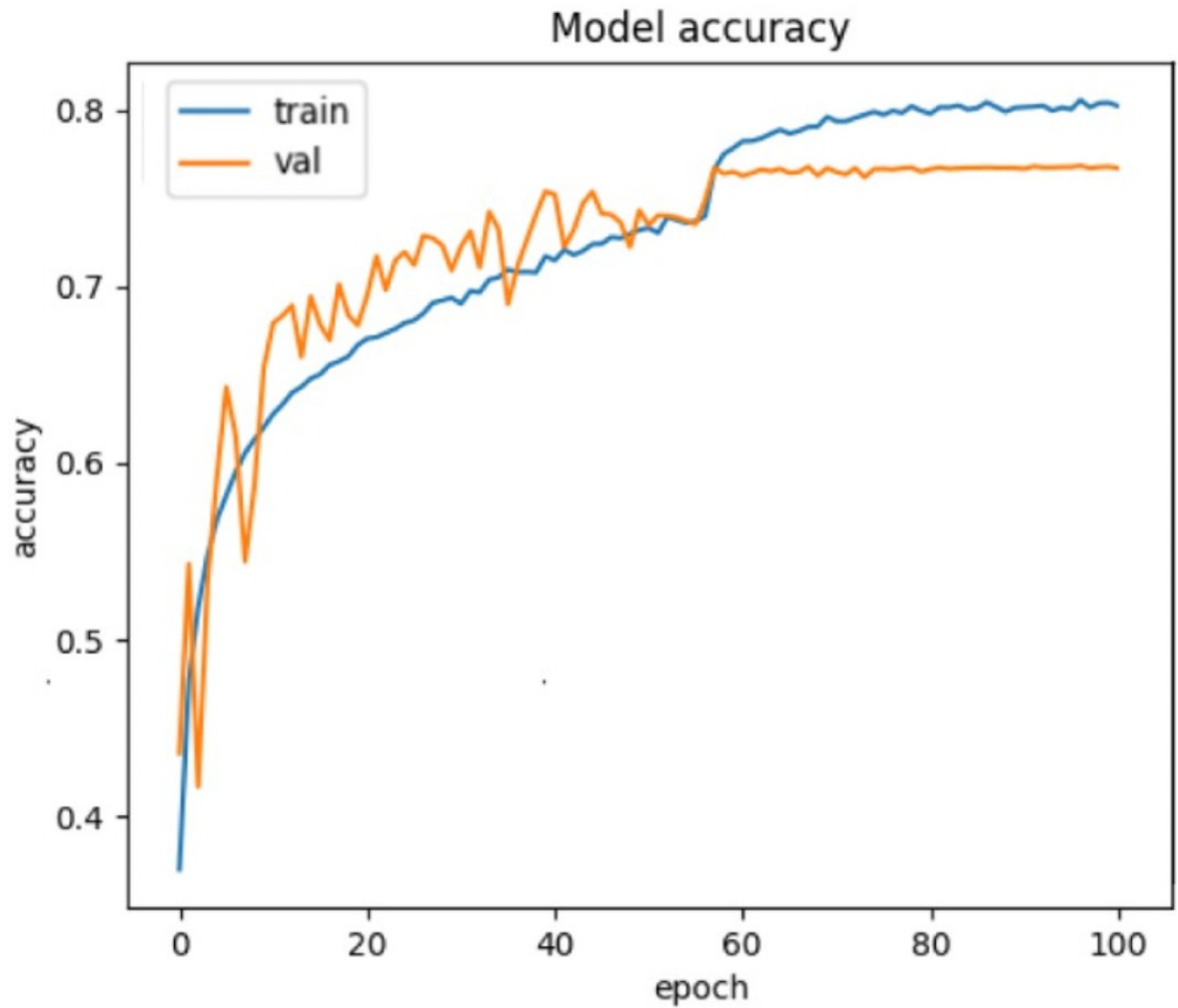

Fig: Accuracy graph for DenseNet-121

As can be seen from the above graphs, maximum accuracy is achieved when DenseNet-121 is used as our model (~76%).

**KEY DIFFERENCES BETWEEN CSE 674 AND CSE 700:**

- We used DenseNet 169 Architecture in addition to DenseNet 121 which was used in CSE 674.
- Results of DenseNet 121 proved to be better than the results given by DenseNet 169.
- In CSE 674, we divided our dataset into 3 phases of 3,330 images each to speed up the evaluation process and make it easier for us to debug. In

CSE 700, we trained our models with the entire dataset of 10,000 images. This improved the accuracy by a considerable amount.
- We trained the three phases of data for 50 epochs in CSE 674. However, in CSE 700, we trained the entire dataset for 100 epochs.

**CONCLUSION:**

COVID-19 has taken millions of lives globally already and still continues to haunt the world as new variants emerge. A common trait among all the variants is their effect on the respiratory system. COVID-19 vaccination became a mandate for all citizens across the world and was given out for free in almost all countries. Despite this, COVID-19 does not seem to end in the near future to say the least. Our project can become handy to many doctors and scientists to at least reduce deaths if not completely avoid.

This comparative examination of the overall performance outcomes of the suggested CNN model provides strong evidence of the model's potential value in COVID-19 diagnosis. The initial dataset had an imbalanced class distribution. This fault in the core dataset had a major influence on the performance of the models. To address this issue, the dataset was preprocessed using a number of techniques. The performance results of the two DenseNet models have been given after training and testing it on the fully processed dataset.

We used two architectures to vividly predict which gives better results. DenseNet-121 stood out considerably with an accuracy of 75.58% whereas DenseNet-169 had an accuracy of 63.25%.

An overall best performance of about 75.58% proves that the DenseNet-121 training approach resulted in enhanced, consistent detection of COVID-19 using RT-PCR standards. While this is far from a perfect measure, it does give a foundation upon which new CNN/DL algorithms might undoubtedly enhance the overall accuracy of the model, allowing for additional RT-PCR funding and application in other current disease detection approaches. As per our research, other algorithms like MobileNet can give better results.